\documentclass[11pt]{article}

\usepackage[preprint]{acl}

\usepackage{times}
\usepackage{latexsym}
\usepackage[T1]{fontenc}
\usepackage[utf8]{inputenc}
\usepackage{microtype}
\usepackage{inconsolata}
\usepackage{graphicx}
\usepackage{booktabs}
\usepackage{multirow}
\usepackage{amsmath}
\usepackage{amssymb}

\title{Answering While You Type:\\Optimizing Time-to-Answer with Streaming Generation}

\author{
  \textbf{Nikita Dragunov},
  \textbf{Matvey Mikhalchuk},
  \textbf{Anton Razzhigaev}
\\
  FusionBrain Lab
\\
  \small{
    \textbf{Correspondence:} \href{mailto:nikitadragunovjob@gmail.com}{nikitadragunovjob@gmail.com}
  }
}

\begin{document}
\maketitle

\begin{abstract}
Work on efficient large-language-model inference has concentrated almost entirely
on reducing the cost of decoding once a request is complete. The quantity a user
experiences, however, is the interval between submitting a request and receiving a
usable response, and that interval can be shortened without making decoding itself
faster. While a request is being composed it is already partially specified, yet
conventional deployments begin no computation until submission. We propose and
study \emph{streaming generation}, a decoding regime in which the model begins
producing a response from an incomplete request and revises it as the remaining
input arrives, so that only a small residue of the answer remains to be computed
once the request is submitted. We evaluate the regime on autoregressive and
diffusion language models, characterise the conditions under which it reduces
user-visible latency, and analyse the mechanism responsible.
\end{abstract}

\section{Introduction}

Research on efficient large-language-model inference optimizes the cost of
producing a response once the request is known, and identifies that cost with the
user's waiting time. This identification is valid only under the convention that decoding begins at
submission. Composing a request takes several seconds, during which the
request is already partially specified and the model performs no work. Streaming
generation uses that interval: the model drafts a response from the partial request
and revises the draft as the remainder arrives, so that little work remains once
the user submits.

The regime changes the quantity to be measured. Total decoding cost ceases to be
the appropriate objective, because computation performed before the user finishes
composing is not part of the observed latency. We therefore measure the
\emph{residual} cost: the forward passes and wall-clock time required after the
final token of the request becomes available.

\paragraph{Contributions.}
\begin{itemize}
\item We formalise streaming generation and adopt residual cost as the evaluation
      criterion, and give a single draft--verify--regenerate procedure that
      instantiates it for both autoregressive and diffusion decoders.
\item We evaluate the regime on six benchmarks spanning the tasks an assistant is
      actually asked to perform: program synthesis from a specification (HumanEval,
      MBPP), data-science code completion (DS-1000), instructed code editing
      (CanItEdit), text editing (CoEdIT) and multi-turn dialogue.
\item On a slot-based diffusion model with an autoregressive verifier
      (ReFusion-8B, \citealp{li2025refusion}) the procedure lowers residual cost by
      $2.18\times$ in forward passes and $1.88\times$ in wall clock, averaged over
      the six benchmarks, without significant loss in quality. The gain is largest
      where the opening of the request already determines the answer, reaching
      $4.67\times$ and $3.68\times$ on HumanEval.
\item On three autoregressive models the same procedure yields macro-averaged median
      residual speedups of $14.4\times$ (Gemma-3-12B), $1.23\times$ (Llama-3.1-8B)
      and $1.21\times$ (Qwen3-8B), at an average quality drop of $2.8$ to $6.1$
      points across the six benchmarks.
\item We show that the procedure does not transfer to fully bidirectional diffusion
      models (LLaDA-8B, Dream-7B), and explain why. Verification there asks the model
      to judge tokens that it can see, which it will almost always approve: offered a
      draft answering a completely different question, it accepts every token. What
      the method needs is a verifier whose judgement of a token is formed without
      access to that token, at a cost that does not grow with the length of the
      answer. Causal attention provides exactly this in a single forward pass.
\item We report that self-distillation on incomplete requests helps, and that what it
      buys differs by model. On Qwen3-8B it roughly doubles the mean residual speedup
      ($11.9\to24.8\times$) and raises the median by a third
      ($1.21\to1.65\times$), by lengthening the drafts the verifier already accepts.
      On Gemma-3-12B, which streams well to begin with, it adds a little of both
      accuracy and speed. On ReFusion-8B it improves accuracy by up to $17$ points.
\end{itemize}

\section{Related Work}

\paragraph{Speculative decoding.}
Speculative decoding accelerates autoregressive generation by having a cheap
drafter propose several tokens that the target model verifies in a single parallel
forward pass, provably preserving the target distribution
\citep{leviathan2023fast,chen2023accelerating}. Follow-up work mostly addresses the
cost and fidelity of the drafter: Medusa attaches lightweight parallel decoding
heads to a frozen backbone \citep{cai2024medusa}, self-speculative decoding drafts
by skipping intermediate layers of the target model itself \citep{zhang2024draft},
and EAGLE autoregresses over hidden features rather than tokens
\citep{li2024eagle}; \citet{xia2024unlocking} survey the area. These methods share
\emph{when} the draft is produced and \emph{from what}: speculation occurs at decode
time, after the request is complete, and the draft originates from a smaller,
early-exiting or auxiliary head. Our setting differs on both counts. The draft is
produced by the same model during composition, so speculative computation occupies
an interval that would otherwise go unused and no auxiliary drafter must be trained
or deployed. Verification at submission serves the function that parallel
verification serves in speculative decoding, but the accepted prefix is carried
across a turn boundary rather than across a few decode steps.

\paragraph{Generation before submission.}
Beginning work before the input is complete is not a new idea, although we are
not aware of prior work in the typed-text setting. Its oldest form is simultaneous
translation, where wait-$k$ policies emit target tokens from a source prefix
\citep{ma2019stacl} and re-translation revises the hypothesis as the source grows
\citep{arivazhagan2020retranslation}; SSBD makes re-translation efficient by reusing
the previous output as a self-speculative draft for the extended input
\citep{zeng2025selfspeculative}. In speech, PredGen performs ``input-time
speculation'', drafting a response from the first ASR chunk and refining it as later
chunks arrive \citep{li2025predgen}; it is the closest neighbour to this work,
differing in modality and in using standard autoregressive speculative decoding.
Outside language, SpeQL begins executing a SQL query before the user has finished
typing it \citep{li2025speql}.

\paragraph{Diffusion language models and ReFusion.}
Discrete diffusion for text was placed on a principled footing by D3PM
\citep{austin2021structured} and later simplified: \citet{sahoo2024simple} show that
a Rao-Blackwellized masked-diffusion objective reduces to a mixture of masked-LM
losses. Scaling the recipe yields general-purpose diffusion LLMs such as LLaDA
\citep{nie2025large} and, through semi-autoregressive and block decoding,
practically deployable variants \citep{han2023ssd,arriola2025block}. The relevant
property is that every position in the current window is re-predicted at each step,
so an already-emitted token may be revised rather than being frozen, as it
necessarily is under autoregressive decoding.

Our principal diffusion backbone is ReFusion-8B \citep{li2025refusion}, which differs
from the models above in two respects that the method depends on. First, its answer buffer is divided into
fixed-size \emph{slots}, and a slot is committed only when the model's confidence in
its contents exceeds a threshold; decoding therefore terminates as soon as the
answer is complete rather than after a fixed number of denoising steps. Second, and critically for this work, its attention is causal and it is trained with
a mixed objective: an
autoregressive loss on committed positions and a masked-diffusion loss on masked
ones. The model consequently exposes two predictors for the same buffer. The
diffusion predictor fills masked slots in place; the autoregressive predictor scores
a token from the state to its left, which cannot observe the token itself. The
second is what makes verification meaningful, and its absence is what we will show
prevents the method from transferring to fully bidirectional diffusion models.
Closest in spirit is SpecDiff \citep{christopher2025speculative}, which uses a
discrete diffusion model as a parallel drafter verified by an autoregressive model;
we likewise pair masked diffusion with an autoregressive verifier, but both roles
are played by a single model and the verified content is carried across turns.

\paragraph{Prefix caching and KV reuse.}
A large body of systems work accelerates inference by caching key-value state across
requests: Prompt Cache precomputes attention states for reusable prompt modules
\citep{gim2024prompt}, SGLang's RadixAttention maintains a radix tree over KV caches
so that shared prefixes are matched automatically \citep{zheng2024sglang},
PagedAttention supplies the memory substrate that makes such sharing practical
\citep{kwon2023efficient}, and CacheBlend relaxes the strict-prefix requirement by
selectively recomputing a small subset of tokens \citep{yao2025cacheblend}. What
unifies them is \emph{what} is reused: the computation belongs to the request, so
they shorten prefill. We reuse the \emph{answer} computed while the user was still
composing, and therefore save decode rather than prefill; the two are composable.

\paragraph{User-perceived latency.}
Latency is conventionally decomposed into time-to-first-token and
time-per-output-token, and systems work optimizes both: Orca introduces
iteration-level scheduling \citep{yu2022orca}, Sarathi-Serve uses chunked prefills
to remove decode stalls \citep{agrawal2024sarathiserve}, and DistServe disaggregates
prefill from decode to meet the two targets independently \citep{zhong2024distserve}.
All of them start the clock at submission, which is the assumption relaxed here.

\section{Streaming Generation}
\label{sec:method}

\paragraph{Modelling the setting.}
The granularity at which a partially composed request becomes available depends on
the application and the domain. A code
assistant may observe every keystroke; a chat interface may see text arrive in
bursts; a voice interface receives transcription chunks whose boundaries are set by
the recogniser. No single segmentation is universally appropriate. We therefore adopt an explicit
simplification: composition is modelled as the arrival of successive \emph{sentences},
and each sentence boundary is treated as a point at which new input becomes
available. This is a controlled proxy for keystroke-level composition, selected
because it is reproducible across benchmarks and because a sentence is the smallest
unit after which a request is generally interpretable.

The simplification rests on one assumption that must be stated. It is meaningful
only if the model can produce and revise a draft faster than the user composes the
next sentence. This condition is satisfied for the models and answer lengths studied
here, since a draft of a few dozen tokens requires a small number of forward passes,
well below the time required to type a sentence. It would have to be re-examined for
considerably longer answers or faster composition.

\paragraph{Cost and speedup.}
A request arrives as a growing sequence of sentence prefixes $S_1 \subset S_2
\subset \dots \subset S_n$, where $S_n$ is the complete request. We report cost in
two units: \emph{forward passes} through the backbone, which is hardware-independent
and is what an implementation can be held to, and \emph{wall-clock seconds}, which
is what the user experiences. A non-streaming system pays the full cost of decoding
$S_n$ after submission. A streaming system performs work at every $S_i$, but only
the work performed after $S_n$ is observable. We therefore report
\[
  \mathrm{speedup} \;=\;
  \frac{c_{\mathrm{base}}(S_n)}{c_{\mathrm{stream}}(S_n \mid D_{n-1})},
\]
where $D_{n-1}$ is the draft carried in from the preceding sentence, and we report
total cost separately so that the compute-for-latency exchange is explicit.

\paragraph{The procedure.}
The same three-step loop instantiates the regime in both model families. The system
holds a draft $D_{i-1}$ produced for the request as it stood at $S_{i-1}$. When
sentence $S_i$ arrives:

\begin{enumerate}
\item \textbf{Verify.} One forward pass over $S_i$ concatenated with $D_{i-1}$
      yields, for every draft position $j$, the probability the model assigns to the
      token already sitting there, conditioned on the updated request.
\item \textbf{Accept.} The longest leading run of draft tokens whose probability
      clears a threshold $\tau$ is retained; the remainder is discarded.
\item \textbf{Regenerate.} Decoding resumes from the accepted prefix and completes
      the answer, producing $D_i$.
\end{enumerate}

At submission the loop runs once more: the draft is verified in a single forward
pass, and only the suffix after the accepted prefix is regenerated. The residual
cost is therefore one verification pass plus the decoding of that suffix, and it is
this quantity, not the cost of a full decode, that the user waits through.

Because the system cannot know which sentence will turn out to be the last, the
procedure is executed at every intermediate point. This is a property of the experimental protocol rather than a deployment
recommendation. A production system would schedule drafting according to load and
available capacity, drafting frequently when the accelerator is underutilised and
not at all when it is saturated. Our measurements assume that spare capacity is
available, which is the regime in which the question arises.

\paragraph{Why the accepted set must be a prefix.}
Step~2 keeps a leading run rather than every token that clears the threshold. For an
autoregressive model the reason is immediate: it writes left to right and cannot
insert text in the middle, so as soon as one token is rejected everything after it
has to be rewritten anyway.

For ReFusion-8B the reason is less obvious, because the model \emph{can} rewrite any
position and so is not forced to keep a prefix. The constraint comes from the
verifier instead. It reports the probability of the token at position $j$ given the
draft tokens to its left, so that number is only meaningful while those tokens stay
in place. Keep a scattered set and every gap invalidates the scores of everything
after it, since the text those scores were conditioned on is about to change. A
leading run is the longest stretch over which the verifier's own numbers still
describe the answer being assembled. We did try the alternative, keeping every token
above threshold and letting the model fill the gaps; it left the answer perforated
throughout rather than truncated at the end, and cost more to repair than it saved.

\paragraph{Family-specific details.}
The two families differ only in step~3. An autoregressive decoder writes strictly
left to right and cannot go back, so it simply resumes writing after the accepted
prefix. A diffusion decoder holds the whole answer in a buffer and can rewrite any
position, so the rejected part is blanked out and filled in again. The practical
difference is that the diffusion model is free to change its mind about text it has
already written, and the autoregressive model is not.

\section{Experimental Setup}

\paragraph{Models.}
We evaluate two decoder families. The autoregressive models are Gemma-3-12B-IT
\citep{team2025gemma3}, Llama-3.1-8B-Instruct \citep{grattafiori2024llama3} and
Qwen3-8B \citep{yang2025qwen3}; drafts are verified by greedy agreement with a newly
decoded continuation under the extended request and truncated at the first
divergence. The diffusion models are ReFusion-8B \citep{li2025refusion}, described in
Section~\ref{sec:method}, and two fully bidirectional models that lack its
autoregressive verification path, LLaDA-8B \citep{nie2025large} and Dream-7B
\citep{ye2025dream}. Decoding is greedy throughout. The diffusion experiments were run on an NVIDIA RTX A6000 and the autoregressive
experiments on NVIDIA A100 80\,GB GPUs.

\paragraph{Benchmarks.}
Both families use the same suite: HumanEval \citep{chen2021humaneval}, MBPP
\citep{austin2021mbpp}, CanItEdit \citep{cassano2024canitedit}, DS-1000
\citep{lai2023ds1000}, CoEdIT \citep{raheja2023coedit} and multi-turn chat drawn
from UltraChat-style dialogues and an MT-Bench-like suite \citep{zheng2023mtbench}.
Requests are split into sentences and replayed cumulatively; requests consisting of a
single sentence are excluded, since there is nothing to stream. For chat, the full
dialogue history is always present and only the final user turn is streamed. Examples
for which either side returns an empty answer are excluded from all columns alike.

\section{Results}
\label{sec:results}

\begin{table*}[t]
\centering
\small
\begin{tabular}{lrrrrrrrr}
\toprule
& \multicolumn{2}{c}{Quality (\%)} & \multicolumn{2}{c}{Forwards after final token} & \multicolumn{2}{c}{Speedup, forwards ($\times$)} & \multicolumn{2}{c}{Speedup, wall clock ($\times$)} \\
\cmidrule(lr){2-3}\cmidrule(lr){4-5}\cmidrule(lr){6-7}\cmidrule(lr){8-9}
Benchmark & base & stream & base & stream & mean & median & mean & median \\
\midrule
HumanEval (pass@1) & 54.3 & 54.3 &  16 &   3 & 5.10 & 4.67 & 3.86 & 3.68 \\
MBPP (pass@1)      & 55.1 & 57.1 &  12 &   3 & 3.76 & 3.33 & 3.21 & 2.90 \\
CanItEdit (pass@1) & 19.7 & 23.7 &  42 &  27 & 3.91 & 1.36 & 2.85 & 1.25 \\
DS-1000 (pass@1)   &  3.9 &  1.6 &  18 &  11 & 2.12 & 1.43 & 1.76 & 1.32 \\
CoEdIT (F1)        & 72.9 & 71.8 &  26 &  19 & 1.82 & 1.07 & 1.63 & 1.03 \\
Chat (F1)          & 42.7 & 42.5 & 194 & 148 & 2.37 & 1.20 & 1.60 & 1.12 \\
\bottomrule
\end{tabular}
\caption{ReFusion-8B. \textbf{base}: decoding the complete request with the
benchmark's default configuration. \textbf{stream}: the procedure of
Section~\ref{sec:method}. Both come from one run over the same examples. Forward
counts are medians. Means exceed medians on every benchmark because the distribution
carries a long right tail (Figure~\ref{fig:refusion-dist}).}
\label{tab:diffusion}
\end{table*}

\begin{table*}[t]
\centering
\small
\setlength{\tabcolsep}{4pt}
\begin{tabular}{llrrrrrr}
\toprule
& & \multicolumn{2}{c}{Quality (\%)} & \multicolumn{2}{c}{Speedup, forwards ($\times$)} & \multicolumn{2}{c}{Speedup, wall clock ($\times$)} \\
\cmidrule(lr){3-4}\cmidrule(lr){5-6}\cmidrule(lr){7-8}
Model & Benchmark & base & stream & mean & median & mean & median \\
\midrule
\multirow{7}{*}{Gemma-3-12B}
             & MBPP (pass@1)      & 75.8 & 65.8 & 75.89 & 74.00 & 63.73 & 67.80 \\
             & HumanEval (pass@1) & 84.1 & 75.0 & 63.05 &  7.98 & 49.13 &  7.67 \\
             & CanItEdit (pass@1) & 41.0 & 30.5 & 53.95 &  1.44 & 20.84 &  1.40 \\
             & DS-1000 (pass@1)   &  0.0 & 10.2 &  2.38 &  1.04 &  2.23 &  1.03 \\
             & CoEdIT (F1)        & 39.0 & 39.2 &  7.89 &  1.05 &  7.07 &  1.05 \\
             & Chat (F1)          & 31.8 & 31.8 &  1.05 &  1.01 &  1.03 &  1.00 \\
             & \textbf{average}   & 45.3 & 42.1 & 34.03 & 14.42 & 24.01 & 13.32 \\
\midrule
\multirow{7}{*}{Llama-3.1-8B}
             & MBPP (pass@1)      & 59.8 & 55.4 & 34.41 &  1.25 & 18.30 &  1.23 \\
             & HumanEval (pass@1) & 67.7 & 62.2 & 34.22 &  2.06 & 19.52 &  2.00 \\
             & CanItEdit (pass@1) & 40.0 & 36.2 &  1.20 &  1.00 &  1.17 &  0.99 \\
             & DS-1000 (pass@1)   &  6.2 &  3.1 &  1.47 &  1.00 &  1.34 &  0.98 \\
             & CoEdIT (F1)        & 51.5 & 51.8 &  1.27 &  1.05 &  1.24 &  1.04 \\
             & Chat (F1)          & 41.0 & 41.0 & 10.18 &  1.01 &  3.29 &  0.99 \\
             & \textbf{average}   & 44.4 & 41.6 & 13.79 &  1.23 &  7.48 &  1.21 \\
\midrule
\multirow{7}{*}{Qwen3-8B}
             & MBPP (pass@1)      & 70.6 & 67.2 & 26.32 &  1.09 & 17.78 &  1.08 \\
             & HumanEval (pass@1) & 83.5 & 52.4 & 25.54 &  2.12 & 20.70 &  2.09 \\
             & CanItEdit (pass@1) & 45.7 & 37.1 &  9.15 &  1.00 &  4.44 &  1.00 \\
             & DS-1000 (pass@1)   & 10.9 & 17.2 &  3.84 &  1.00 &  2.81 &  0.98 \\
             & CoEdIT (F1)        & 59.2 & 59.2 &  5.51 &  1.02 &  4.67 &  1.02 \\
             & Chat (F1)          & 36.7 & 36.7 &  1.05 &  1.01 &  1.03 &  1.00 \\
             & \textbf{average}   & 51.1 & 45.0 & 11.90 &  1.21 &  8.57 &  1.19 \\
\bottomrule
\end{tabular}
\caption{Autoregressive models with no adaptation, per benchmark. \emph{average} is
the macro-average over the six benchmarks. The gain concentrates on the two
specification-first coding benchmarks, as it does for the diffusion model in
Table~\ref{tab:diffusion}, and means run far above medians for the same reason.}
\label{tab:ar}
\end{table*}

On ReFusion-8B the procedure reduces residual cost substantially on the two
specification-first coding benchmarks and leaves accuracy intact: HumanEval is
$4.67\times$ cheaper in forward passes and $3.68\times$ faster in wall clock at
unchanged pass@1, and MBPP is $3.33\times$ and $2.90\times$ with pass@1 rising by two
points (Table~\ref{tab:diffusion}). The remaining four benchmarks gain between
$1.07\times$ and $1.43\times$.

The procedure transfers to the autoregressive family on all three checkpoints, with
the same task dependence and a wider spread in magnitude (Table~\ref{tab:ar}).
Gemma-3-12B streams exceptionally well with no adaptation at all, reaching a median
residual speedup of $14.4\times$. Llama-3.1-8B and Qwen3-8B return medians of
$1.23\times$ and $1.21\times$, close to a fifth of the post-submission work removed
on the typical request, together with means of $13.8\times$ and $11.9\times$ carried
by a substantial minority of requests that finish almost immediately. All three share
the same acceptance rule, so the spread is a property of the individual checkpoint
rather than of the decoder family. Streaming costs $2.8$ to $6.1$ accuracy points,
concentrated on the coding benchmarks.

\section{Analysis}
\label{sec:analysis}

\begin{figure*}[t]
\centering
\includegraphics[width=\textwidth]{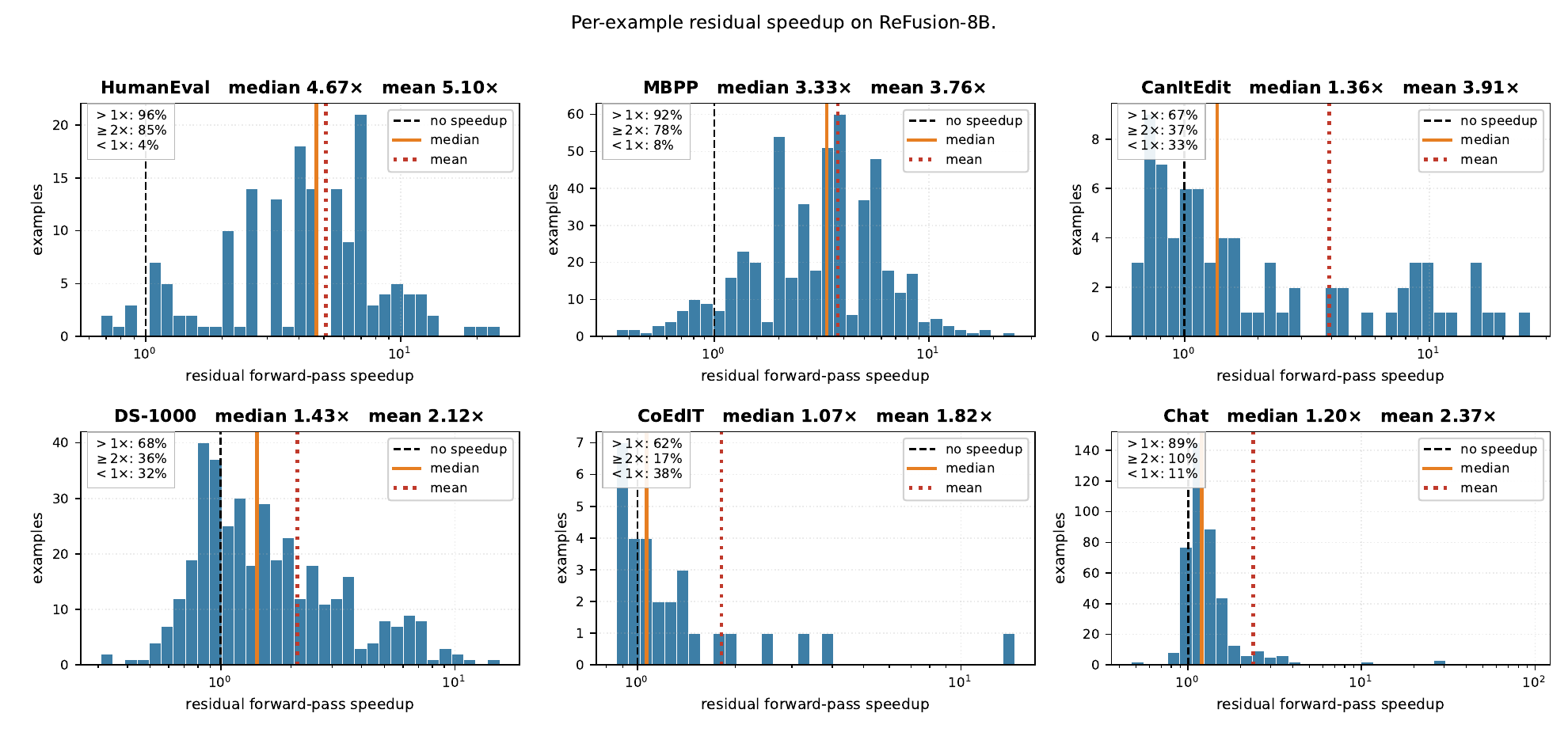}
\caption{Per-example residual speedup on ReFusion-8B, one panel per benchmark, on a
logarithmic axis. Orange marks the median, dotted red the mean, dashed black the $1\times$ line at which streaming
costs what the baseline costs.}
\label{fig:refusion-dist}
\end{figure*}

\subsection{The distribution of the gain}

Figure~\ref{fig:refusion-dist} shows the per-example residual speedup on ReFusion-8B,
one panel per benchmark. The distributions have three properties that a single
summary statistic would not convey, and each of them bears on how the method should
be deployed.

The first is that the gain varies widely between requests of the same benchmark. On
HumanEval and MBPP most requests fall between $2\times$ and $8\times$, and individual
requests reach $24\times$. On CanItEdit the most common outcome is close to
$1\times$, yet a minority of requests exceeds $10\times$. Because the distributions
are right-skewed, the mean and the median answer different questions: the mean
reflects the total computation saved across a workload, while the median describes
the request a typical user issues. Tables~\ref{tab:diffusion} and~\ref{tab:ar}
therefore report both.

The second is that every benchmark contains requests that gain nothing. Between $4\%$
and $38\%$ of them fall below $1\times$. This follows from the design of the
procedure rather than from any defect in it: one verification pass is always spent,
and when the draft is rejected that pass yields nothing, leaving the request slightly
more expensive than the non-streaming baseline.

The third is that the outcome is close to binary rather than graded, and this is what
gives the distributions their shape. At the final prompt update $42\%$ of requests
keep no draft token at all and $46\%$ keep every one, leaving only $12\%$ in between.
Requests that keep nothing are concentrated just below $1\times$, and requests that keep
their draft account for the entire right tail. Benchmarks therefore differ chiefly in
how often the draft is kept, from $9\%$ of HumanEval and MBPP requests discarding it
to $56\%$ on DS-1000 and $82\%$ on chat. Partial acceptance is uncommon, though more
frequent where the request is an instruction to edit existing text: it occurs on
$30\%$ of CanItEdit requests against $2\%$ on MBPP.

\begin{figure*}[t]
\centering
\includegraphics[width=\textwidth]{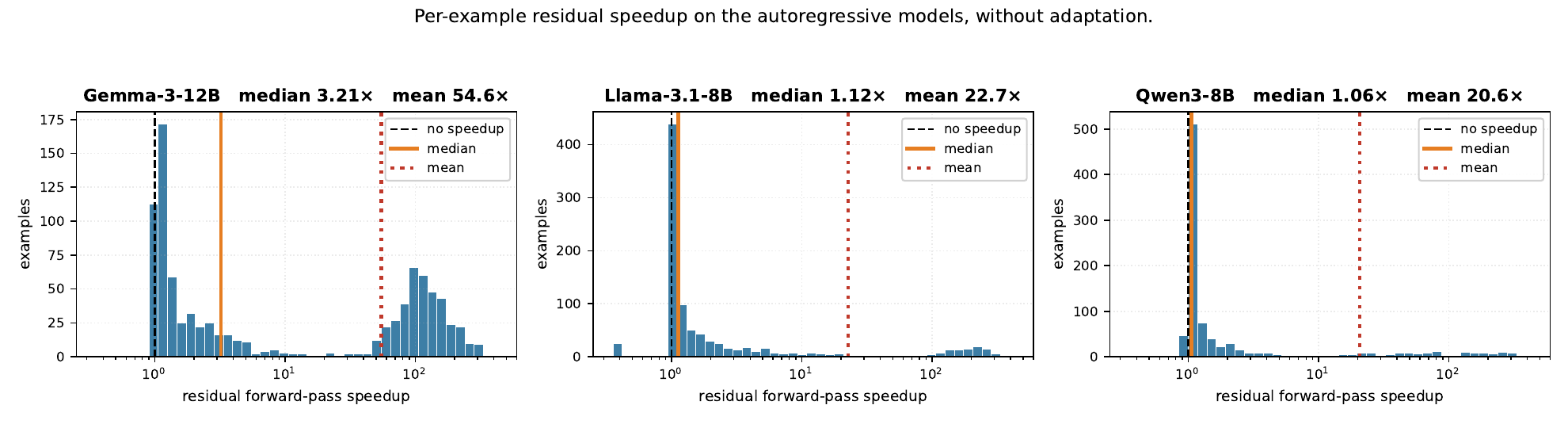}
\caption{Per-example residual speedup on the three autoregressive models, on the same
axes as Figure~\ref{fig:refusion-dist}. Orange marks the pooled median, dotted red the
pooled mean, dashed black the $1\times$ line.}
\label{fig:ar-dist}
\end{figure*}

The autoregressive models show the same skew, in a more extreme form
(Figure~\ref{fig:ar-dist}). All three place a large spike at $1\times$ and a long
tail beyond it, so that pooled medians of $3.21\times$, $1.12\times$ and $1.06\times$
sit alongside pooled means of $54.6\times$, $22.7\times$ and $20.6\times$. Gemma-3-12B differs from the other two in having a
distinct second mode above $100\times$: on requests whose opening determines the
answer it accepts drafts of several hundred tokens and finishes $34\%$ of requests in
a single residual forward pass, against roughly $10\%$ for Llama and Qwen. Reporting
the mean alone on this family would describe a request that almost never occurs. The
medians quoted here are taken over the pooled distribution of individual requests,
whereas Table~\ref{tab:ar} macro-averages each benchmark's own median.

\subsection{Streaming is worth applying unconditionally}

Taken together, these distributions support a simple deployment policy: stream every
request, without attempting to predict in advance which ones will benefit.

The expected gain is substantial. Pooled over the $1573$ diffusion requests, the mean
residual speedup is $3.10\times$ and the median $1.68\times$, and streaming removes
$37.6\%$ of the forward passes that the user would otherwise wait through. Both
averages favour the method, so it is not the case that a rare success is being paid
for by a common loss.

The risk is correspondingly small, and this is the more important half of the
argument. Only $16\%$ of requests end below $1\times$, $10\%$ below $0.9\times$ and
$5\%$ below $0.8\times$. Among the requests that do lose, the median outcome is
$0.86\times$, that is a delay of roughly one seventh of the baseline cost. A
pronounced slowdown is rare because the penalty for a rejected draft is structurally
bounded: it is one wasted verification pass, whose cost does not grow with the length
of the answer, whereas the reward for an accepted draft grows with it. The asymmetry
between a bounded loss and an unbounded gain is what makes the unconditional policy
preferable, and it also means that a gate would have to be very accurate to be worth
its own complexity.

\subsection{Where it works, and why}

Streaming pays off when the request is of the form ``produce an artefact from a
specification''. The specification appears in the opening sentences, and what follows
is illustration (examples, assertions, formatting markers) that refines the answer
without changing its structure. The draft written from the first sentence is
then already close to the final answer, it survives each update, and at submission
only a verification pass and a short remainder are left. HumanEval and MBPP are of
this form, and both families behave the same way on them: the diffusion model reaches
$4.67\times$ and $3.33\times$, and the strongest autoregressive checkpoint reaches
$74\times$ on MBPP.

Two things reduce the gain. One is late-arriving instruction: in editing tasks the
program is supplied first and the instruction last, so the model correctly writes
nothing until the final sentence and there is no draft to reuse. The other is the
absence of a unique correct wording. In open dialogue a semantically equivalent
rephrasing fails a token-level acceptance test, so drafts are discarded even when they
were perfectly good answers. Chat sits at the bottom of the table for this reason
rather than because its requests are hard.

There is also an upper limit on what the method can achieve, which is worth stating
because the observed values approach it. Even when the entire draft is accepted, the
model must still perform a minimum number of forward passes to confirm the answer and
emit it; on ReFusion-8B that minimum is three, and no request in the evaluation
completed in fewer. The attainable speedup on a request is therefore its baseline
cost divided by three, which for the median request is $5.33\times$ on HumanEval and
$4.00\times$ on MBPP. The measured $4.67\times$ and $3.33\times$ correspond to $88\%$
and $83\%$ of those limits, so on these benchmarks little further gain is available
from accepting more draft tokens.

\subsection{Why the two families differ in what they pay}

The diffusion and autoregressive results differ in one respect that the tables make
plain: ReFusion-8B preserves quality while the autoregressive checkpoints do not.
Table~\ref{tab:ar} locates the loss precisely. On CoEdIT and chat the quality change
is between $0.0$ and $+0.3$ points for all three models, and on the code and editing
benchmarks it reaches $-31$. Those are also the two benchmarks on which the
autoregressive median speedup is $1.01$--$1.05\times$, that is, on which almost no
draft is accepted.

The two observations are the same observation. An autoregressive decoder cannot
revise a token it has emitted, so whatever part of the draft is accepted is fixed for
good, including the part that was written before the request was complete. Where
nothing is accepted the answer is the non-streaming answer and quality is untouched;
where a long prefix is accepted the model is committed to an opening chosen under
less information than it now has. The diffusion decoder is not subject to this,
because it re-predicts every position in its buffer and can revise an accepted
region if the completed request calls for it. The quality cost of streaming is
therefore not intrinsic to the regime. It is the price of irreversible decoding, and
it is paid only by the family that decodes irreversibly.

\subsection{Where it does not work}

The procedure needs a verifier that can be wrong about the draft: one whose judgement
of a token is formed without access to that token. Causal attention supplies this for
free, since the state that scores position $j$ has seen only what lies to its left,
and one forward pass yields a verdict on the entire draft.

Fully bidirectional diffusion models supply nothing of the kind, and the consequence
is not subtle. Offered a draft answering an entirely unrelated request, an account of
the Byzantine Empire presented as the answer to a Python exercise, Dream-7B assigns
it a median probability of $0.999$ and accepts every one of its $96$ tokens.
It is not indiscriminate: random tokens are rejected everywhere. It is confirming
that the text in front of it is the text in front of it. Applied to LLaDA-8B and
Dream-7B the procedure consequently keeps stale drafts and loses accuracy at every
threshold we tried; on HumanEval nearly every streamed Dream answer is the same
nine-token stub, written from the opening sentence and never revised. Hiding the
draft to force an honest reading does not help, because it also removes the context
the draft needs in order to be judged at all.

\section{Adapting the Model to Incomplete Requests}
\label{sec:adaptation}

Streaming asks a model to do something it was not trained for: answer a request that
stops mid-sentence, and then agree with itself once the remainder arrives. We
therefore adapted models in both families to that regime.

\paragraph{Setups.}
Supervision is the answer to the \emph{complete} question given only a prefix of it,
so that the model learns to commit early to what it will eventually say. Adaptation
uses LoRA \citep{hu2022lora} on the attention projections, and autoregressive runs
additionally prepend a system message asking the model to answer incomplete
questions, following \citet{li2025predgen}. The autoregressive adapters use rank
$8$, $\alpha{=}16$, dropout $0.05$, learning rate $2\times10^{-5}$, effective batch
size $16$ and a maximum sequence length of $2048$; evaluation uses the merged
checkpoint at $1000$ steps.

The two families were adapted on different corpora, so we report each separately and
draw no cross-family comparison. For ReFusion-8B we built $13{,}845$
prefix-to-answer pairs from three code corpora disjoint from the evaluation suite:
the MBPP training split \citep{austin2021mbpp}, Magicoder-OSS-Instruct
\citep{wei2024magicoder} and Evol-CodeAlpaca \citep{luo2024wizardcoder}. The
autoregressive models were adapted on requests drawn from LMSYS-Chat-1M
\citep{zheng2024lmsyschat}, a corpus of real user conversations with deployed
assistants. The two adaptation corpora therefore differ in domain as well as in
construction, which is the reason we do not compare the families on the effect of
adaptation.

For the diffusion model we trained two variants: \emph{draft-refinement}, which is
shown the current draft and learns to repair it, and \emph{prefix-to-answer}, which
maps a bare request prefix to the full answer with no draft in the input.

\begin{table}[t]
\centering
\small
\setlength{\tabcolsep}{4pt}
\resizebox{\columnwidth}{!}{%
\begin{tabular}{@{}llcc@{}}
\toprule
Model & Benchmark & quality & median speedup \\
\midrule
\multicolumn{4}{l}{\emph{Autoregressive}} \\
\multirow{2}{*}{Qwen3-8B}
             & MBPP      & $67.2\to59.2$ & $ 1.09\to 1.50$ \\
             & HumanEval & $52.4\to59.8$ & $ 2.12\to 3.44$ \\
Llama-3.1-8B & average   & $41.6\to41.7$ & $ 1.23\to         1.18 $ \\
Gemma-3-12B  & average   & $42.1\to43.3$ & $14.42\to        15.20 $ \\
\midrule
\multicolumn{4}{l}{\emph{Diffusion (ReFusion-8B)}} \\
\multirow{2}{*}{draft-refinement}
             & HumanEval & $54.3\to57.3$ & $ 4.67\to         4.67 $ \\
             & MBPP      & $57.1\to59.1$ & $ 3.33\to         2.93 $ \\
\multirow{2}{*}{prefix-to-answer}
             & HumanEval & $54.3\to62.2$ & $ 4.67\to         3.56 $ \\
             & MBPP      & $57.1\to53.4$ & $ 3.33\to         2.13 $ \\
\bottomrule
\end{tabular}}
\caption{Adaptation to incomplete requests, before~$\to$~after.}
\label{tab:adaptation}
\end{table}

\paragraph{What improves.}
Adaptation helps, and what it buys depends on the model
(Table~\ref{tab:adaptation}). Qwen3-8B gains the most speed: the macro-averaged mean
residual forward speedup roughly doubles ($11.9\to24.8\times$, wall clock
$8.6\to15.3\times$) and the median rises on every benchmark we measured
(MBPP $1.09\to1.50\times$, HumanEval $2.12\to3.44\times$, CanItEdit
$1.00\to1.52\times$, CoEdIT $1.02\to1.39\times$). The mechanism is acceptance length:
the fraction of draft tokens preserved rises from $0.27$ to $0.43$, and the share of
requests finishing in a single residual forward pass from $9.6\%$ to $15.9\%$. Gemma
gains a little of both quality and speed, Llama gains in the mean.

On ReFusion-8B adaptation buys quality rather than speed. Every benchmark improves,
several substantially: $+3.0$ points on HumanEval, $+2.0$ on MBPP, $+0.8$ on DS-1000,
$+7.7$ on CanItEdit and $+17.1$ on CoEdIT. The prefix-to-answer variant reaches the
highest HumanEval quality of any configuration we ran, $+7.9$ points, but is less
stable across benchmarks. Median speed is unchanged on HumanEval and slightly lower
elsewhere.

The pattern is consistent with the analysis above. Adaptation makes the generator
commit earlier, which lengthens drafts that the verifier was already accepting; it
does not change whether a draft clears the acceptance test in the first place, and
that is what sets the residual cost of a typical request.

\section{Conclusion}

Time to answer, rather than decoding latency, is what a user experiences, and it can
be reduced without making decoding faster. Streaming generation does so on both
decoder families we studied. On the diffusion model it removes $37.6\%$ of the forward passes a user waits
through. On the two benchmarks whose opening already determines the answer it is
$4.67\times$ and $3.33\times$ cheaper in forward passes and $3.68\times$ and
$2.90\times$ faster in wall clock, at quality that is unchanged on HumanEval and two
points higher on MBPP. On the autoregressive side the gain is larger still where the
model commits early: Gemma-3-12B reaches a median of $74\times$ on MBPP and finishes
a third of all requests in a single forward pass after submission. Adapting a model
to incomplete requests raises the gain further, roughly doubling the mean on
Qwen3-8B.

Two properties decide whether a model can do this. Its verifier must be able to
reject a stale draft, which requires that a token be judged without being seen;
causal attention supplies this in one forward pass, and full bidirectional attention
does not supply it at all. And because a rejected draft costs one verification pass
while an accepted one saves work in proportion to the length of the answer, streaming
is worth applying to every request rather than to a selected few. The computation it
spends is computation that would otherwise sit idle while the user is still typing.

\section*{Limitations}

\paragraph{Latency is bought with computation.}
The regime trades throughput for the latency of an individual user, and the exchange
rate should be stated. Streaming regenerates the answer at every sentence, so total
forward passes rise by a factor of $1.3$ to $5.4$ across the diffusion benchmarks,
and by $3.5$ to $3.9$ on the two where the residual gain is largest. The
autoregressive side likewise performs speculative work during composition
that residual metrics do not charge. This work is done while the accelerator would
otherwise be idle, so the method suits interactive or local deployment and is not
intended for a saturated batch system.

\paragraph{Speed does not imply quality.}
A response may be produced quickly because it is deficient. Among the $44$ diffusion
chat examples with speedup at least $2\times$, $10$ lose a correct answer; one
CanItEdit example reaches $25\times$ while turning correct into incorrect, because a
frozen draft ignored the edit instruction entirely. On the autoregressive side Qwen3-8B
loses $31$ points on HumanEval under streaming.

\paragraph{Variation across checkpoints.}
The three autoregressive models use the same acceptance rule, yet their median
residual speedups range from $1.2\times$ to $14\times$. The one that streams best is
also the largest, but three checkpoints are not enough to tell whether size is the
reason. Why one instruction-tuned model settles on its answer early, while another
keeps revising it, is left open.

\bibliography{custom}

\end{document}